\documentclass[10pt,twocolumn,letterpaper]{article}

\usepackage{wacv}      

\usepackage{graphicx}
\usepackage{amsmath}
\usepackage{amssymb}
\usepackage{booktabs}
\usepackage{multirow}
\usepackage{bbold}

\usepackage[pagebackref,breaklinks,colorlinks]{hyperref}

\usepackage[capitalize]{cleveref}
\crefname{section}{Sec.}{Secs.}
\Crefname{section}{Section}{Sections}
\Crefname{table}{Table}{Tables}
\crefname{table}{Tab.}{Tabs.}

\def\wacvPaperID{16} 
\def\confName{WACV}
\def\confYear{2025}

\begin{document}

\title{Diffusion-Guided Gaussian Splatting for Large-Scale Unconstrained 3D Reconstruction and Novel View Synthesis}

\author{{Niluthpol Chowdhury Mithun$^{1*}$,~ Tuan Pham$^{2}$}\thanks{\textsuperscript{}Equal Contribution},~ Qiao Wang$^{1}$,~Ben Southall$^{1}$,~Kshitij Minhas$^{1}$,\\~Bogdan Matei$^{1}$,~Stephan Mandt$^{2}$,  ~Supun Samarasekera$^{1}$,~Rakesh Kumar$^{1}$ \and
{\normalsize $^{1}$SRI International, Princeton, NJ, USA} \and {\normalsize $^{2}$University of California, Irvine, CA, USA} 
\and
\texttt{\small $^{1}$firstname.lastname@sri.com} \and \texttt{\small $^{2}$\{tuan.pham;~mandt\}@uci.edu}
}
\maketitle

\begin{abstract}

Recent advancements in 3D Gaussian Splatting (3DGS) and Neural Radiance Fields (NeRF) have achieved impressive results in real-time 3D reconstruction and novel view synthesis. However, these methods struggle in large-scale, unconstrained environments where sparse and uneven input coverage, transient occlusions, appearance variability, and inconsistent camera settings lead to degraded quality. We propose GS-Diff, a novel 3DGS framework guided by a multi-view diffusion model to address these limitations. By generating pseudo-observations conditioned on multi-view inputs, our method transforms under-constrained 3D reconstruction problems into well-posed ones, enabling robust optimization even with sparse data. GS-Diff further integrates several enhancements, including appearance embedding, monocular depth priors, dynamic object modeling, anisotropy regularization, and advanced rasterization techniques, to tackle geometric and photometric challenges in real-world settings. Experiments on four benchmarks demonstrate that GS-Diff consistently outperforms state-of-the-art baselines by significant margins.




\end{abstract}

\section{Introduction}
\label{sec:intro}


Recent progress 3DGS and NeRFs have revolutionized real-time 3D reconstruction and view synthesis, particularly under controlled conditions with dense inputs~\cite{kerbl20233d, mildenhall2021nerf}. However, these methods falter in unconstrained environments where sparse data, transient occlusions, appearance variability, varying camera models, and image acquisition issues introduce significant artifacts and quality degradation. While recent works have sought to address specific challenges with tailored modules~\cite{xiong2023sparsegs, kulhanek2024wildgaussians, charatan2024pixelsplat}, most methods remain optimized for small-scale benchmarks, limiting their generalization to unconstrained real-world scenarios.



\begin{figure}
  \centering
\includegraphics[width=0.95\linewidth]{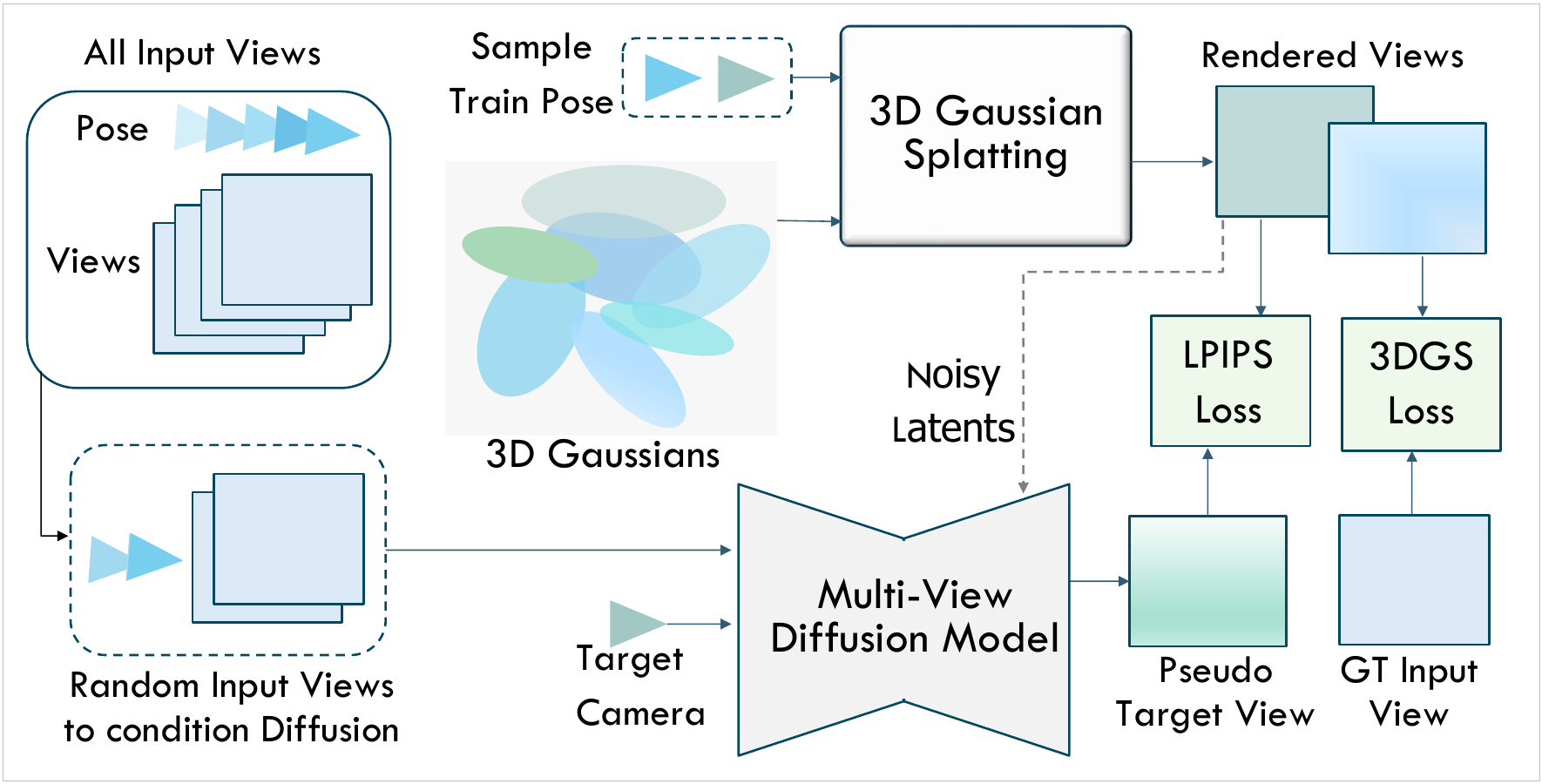}
\vspace{-0.2cm}
  \caption{Brief Illustration of the Proposed GS-Diff Approach.}
  \label{fig:overall}
\vspace{-0.4cm}
\end{figure}

In such environments, sparse and uneven input coverage creates an under-determined reconstruction problem, leading to incomplete geometry and subpar rendering. To address this, we introduce GS-Diff, a novel adaptation of the 3DGS framework guided by prior knowledge from multi-view diffusion model (Fig.~\ref{fig:overall}). 
GS-Diff synthesizes pseudo-observations conditioned on input images using the diffusion model, which provides supplementary viewpoints for 3DGS optimization, effectively converting the 3D reconstruction problem into a more constrained setting. These synthesized views enhance the optimization pipeline, significantly improving reconstruction fidelity and novel view synthesis under sparse and inconsistent conditions.


To further address the challenges unique to unconstrained in-the-wild scenarios, GS-Diff incorporates several enhancements into the 3DGS framework. These include monocular depth priors for improved geometric constraints, appearance embeddings to account for illumination variability, 
dynamic object modeling to address transient occlusions, anisotropy regularization to prevent over-elongated Gaussians, and advanced rasterization techniques such as mip-filtering and absolute gradients to reduce aliasing and blurring. Together, these adaptations make GS-Diff robust to real-world complexities.


Experiments on diverse datasets~\cite{snavely2006photo, cjk5-gf33-24, 2zs6-ht63-24, ren2024nerf}, including ULTRRA benchmark, demonstrate GS-Diff achieves significant improvements over existing methods, bridging the gap between controlled benchmarks and real-world applications for scalable, high-fidelity 3D scene reconstruction.


\section{Related Works}
\label{sec:related}


\textbf{3D Gaussian Splatting for unconstrained scenes: }
3DGS-based methods have recently demonstrated significant advancements in 3D scene reconstruction, particularly in controlled benchmarks~\cite{kerbl20233d,lu2024scaffold, kheradmand20243d}. Several recent works have improved vanilla 3DGS method to tackle specific challenges in 3D reconstruction, such as improving rendering quality and fine-grained details~\cite{yu2024mip, ye2024absgs, yang2024spec, yu2024gaussian}, addressing sparse-view reconstruction~\cite{xiong2023sparsegs, zhu2025fsgs, fan2024instantsplat}, handling lighting variations~\cite{dahmani2025swag, kulhanek2024wildgaussians}, transient occlusions~\cite{xu2024wild, kulhanek2024wildgaussians}, and mitigating camera artifacts~\cite{peng2025bags}. These methods have broadened the applicability of 3DGS in structured scenarios with prior knowledge of scene characteristics. Regression-based generalizable 3DGS methods have also emerged, that directly predict 3D representations from a small number of input images using feed-forward models~\cite{charatan2024pixelsplat, xu2024depthsplat, chen2025mvsplat}. These approaches bypass time-consuming optimization steps, but they often focus mainly on sparse-view cases and fail to produce high-quality view synthesis when exposed to out-of-distribution inputs. However, in real-world, unconstrained scenes, we frequently lack any prior knowledge of the environment or the types of challenges that might arise. It compounds the difficulty of applying current methods effectively, as they are often ill-equipped to handle unexpected complexities or variations in these settings.

\textbf{Multi-View Diffusion for Novel View Synthesis: }
Recent advancements in diffusion models have highlighted their potential for synthesizing novel views, particularly in scenarios with sparse datasets. Multi-view diffusion models such as EscherNet~\cite{kong2024eschernet}, Cat3D~\cite{gao2024cat3d}, and ViewCrafter~\cite{yu2024viewcrafter} have shown decent results in generating plausible views from limited reference images. These models leverage powerful generative priors to extrapolate missing details and enhance visual coherence, offering an alternative to traditional neural rendering pipelines. Despite their promise, these models are typically constrained to small spatial regions with limited reference images, making them prone to hallucinations in complex scenes. Moreover, their outputs often lack strict 3D consistency, posing challenges for direct integration into reconstruction pipelines. 
While few recent efforts have explored ways to integrate diffusion models within reconstruction frameworks~\cite{gao2024cat3d, wu2024reconfusion, liu2024reconx}, challenges remain in achieving 3D consistency, scaling to large scenes, and generalizing to diverse real-world conditions.



\section{Proposed Approach}
\label{sec:approach}


\subsection{3D Gaussian Splatting Baseline}
 \textbf{3D Gaussian Splatting: } 3DGS~\cite{kerbl20233d} represents a scene as trainable 3D Gaussians, each defined by a center \( \mu \in \mathbb{R}^3 \), covariance matrix \( \Sigma \), opacity \( \alpha \), and spherical harmonic (SH) coefficients for view-dependent color \( c \). These Gaussians are projected onto the image plane~\cite{zwicker2001ewa}, yielding 2D Gaussians with transformed means and covariances. A tile-based rasterizer sorts the Gaussians by depth and computes final pixel colors \( \hat{C} \) using \(\alpha\)-blending:  
\[
\hat{C} = \sum_{i=1}^n c_i \alpha_i \prod_{j=1}^{i-1} (1 - \alpha_j)
\tag{1}
\]
where \( c_i \) is SH-based color and \( \alpha_i \) is the opacity-weighted contribution of the \( i \)-th splatted Gaussian. With a given set of images with known poses, training optimizes SSIM and \( L_1 \) losses between predicted image \( \hat{I} \) and ground truth \( I \):  
\[
\mathcal{L}_{\text{GS}} = \lambda_{\text{ssim}} \cdot \text{SSIM}(\hat{I}, I) + (1 - \lambda_{\text{ssim}}) \cdot \|\hat{I} - I\|_1
\tag{2}
\]  


During training, Gaussians with low opacity or large size are pruned, while those with high gradients are split or cloned heuristically to improve 3D representation.

\textbf{Scaffold-GS: } 3DGS's heuristic splitting and cloning often leads to Gaussian drift and redundancy. Scaffold-GS~\cite{lu2024scaffold} addresses this by organizing 3D Gaussians around anchor points derived from Structure-from-Motion (SfM). Each anchor is linked to a feature vector encoding local scene structure and generates \( k \) neural Gaussians with positions \( \mu_i = x_v + O_i \cdot l_v \), where , \( x_v \) is anchor position, \( O_i \) are predicted offsets and \( l_v \) is a scaling factor.  

Neural Gaussian properties like opacity, scale, rotation, and color are decoded from anchor features via multi-layer perceptrons (MLPs). For instance, opacity \( \alpha_i \) is computed as \( \alpha_i = F_\alpha(\hat{f}_v, \Delta v_c, \vec{d}_{v_c}) \), using an MLP \( F_\alpha \), anchor feature \( \hat{f}_v \), viewing distance \( \Delta v_c \), and camera direction \( \vec{d}_{v_c} \).  

During densification, new anchors are added where Gaussian gradients are high, and low-transparency anchors are pruned, improving robustness and storage efficiency compared to vanilla 3DGS. We adopt Scaffold-GS as our base 3D reconstruction pipeline.







\begin{figure}[t]
  \centering
\includegraphics[width=0.92\linewidth]{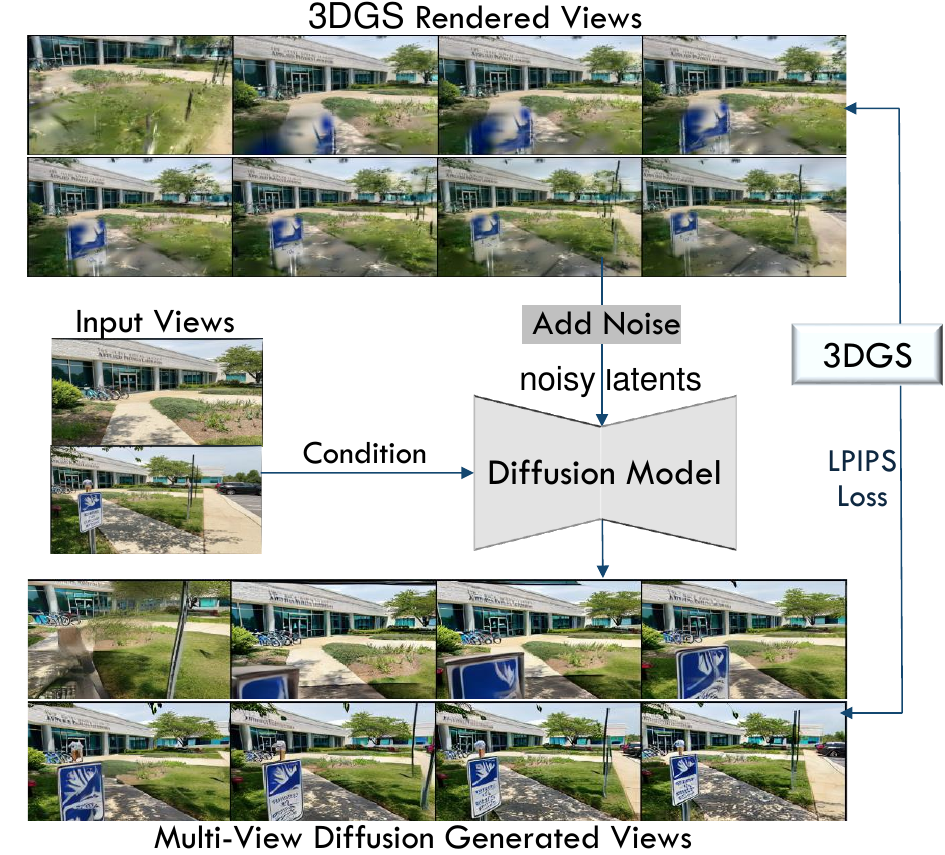}
\vspace{-0.2cm}
  \caption{Iterative workflow of the integrated diffusion process with the proposed GS-Diff pipeline.}
  \label{fig:diffusion}
  \vspace{-0.4cm}
\end{figure}

\subsection{Incorporating Diffusion Priors}


3D reconstruction with limited inputs in unconstrained scenes often leads to overfitting or degraded performance at novel viewpoints. To address this, we integrate prior knowledge in the GS training via a diffusion model (trained for multi-view consistent view synthesis) (Fig.~\ref{fig:overall}).

\textbf{Multi-View Diffusion Prior in 3DGS Training: } The Multi-View (MV) Diffusion model predicts scene appearance from novel viewpoints while maintaining consistency with given posed images. Pre-trained on public multi-view datasets, the model regularizes 3DGS by generating augmented pseudo-training views conditioned on nearby views. An additional loss aligns diffusion-augmented views with corresponding 3DGS-rendered views. Generating all the augmented views one time makes the model prone to hallucinations and 3D inconsistencies of the diffusion model. In this regard, we develop an integrated framework that applies diffusion-based view augmentation iteratively at every $N^{th}$ step during the 3DGS training~(Fig.~\ref{fig:diffusion}).


Training pairs with proximal camera poses are selected, and a spline interpolation generates intermediate camera trajectories. Target augmented cameras sample new views, where noisy 3DGS-rendered latents are input to the MV Diffusion model. The diffusion model denoises these inputs into high-quality generated images (${I_{i}^{D}}$), and compared to 3DGS-rendered views ($\hat{I}_i$) using LPIPS loss~\cite{zhang2018unreasonable, wu2024reconfusion}. LPIPS prioritizes high-level semantic similarity while potentially overlooking low-level inconsistencies, making it suitable for ours. We define LPIPS-based loss as:  
\[
L_{\text{GS}}^{\text{D}} = \boldsymbol{\mathbb{1}}\left(L_{\text{LPIPS}}(\hat{I}_i, I_{i}^{D}) \leq \epsilon\right) \cdot L_{\text{LPIPS}}(\hat{I}_i, {I_{i}^{D}})
\tag{3}
\]
It excludes rendered-generated pairs with LPIPS exceeding $\epsilon$ (e.g., \(\epsilon = 0.5\)) from optimization to ensure stable training.




\textbf{Diffusion Model Training: }EscherNet serves as our baseline diffusion model~\cite{kong2024eschernet}. EscherNet leverages a pre-trained 2D diffusion model (Stable Diffusion), and augments it with a camera positional encoding to handle arbitrary numbers of reference and target views. Its lightweight view encoding and scene-agnostic design, which avoids volumetric operations or ground-truth geometry, make it both efficient and adaptable to incorporate in our pipeline.

To align EscherNet effectively with the requirements of efficient 3DGS optimization, we introduce two key modifications in training: (1) training on lower-resolution images with fewer intermediate views to reduce computational overhead, and (2) employing a noise schedule shift towards higher noise levels, similar to~\cite{gao2024cat3d}. These adaptations enable EscherNet to complement 3DGS effectively, particularly under challenging real-world conditions.


\begin{table}[t]
\centering
\renewcommand{\arraystretch}{1.05}
\setlength{\tabcolsep}{3 pt}
\caption{DreamSim Comparison on ULTRRA and WRIVA Sets.}
\vspace{-0.2cm}
\resizebox{\linewidth}{!}{%
\begin{tabular}{l|cc|cccc|cc}
\toprule
\multirow{3}{*}{\shortstack{Dataset \\ (no. images)}} & \multicolumn{2}{c|}{ULTRRA} & \multicolumn{4}{c|}{WRIVA-AIDI} & \multicolumn{2}{c}{WRIVA-MTA} \\
\cmidrule(lr){2-3} \cmidrule(lr){4-7} \cmidrule(lr){8-9}
 & ID-1901 & CM-2601 & S05 & S06 & S07 & S08 & S01 & S02 \\
  & (96)  & (77) & (25) & (15) & (10) & (5) & (50) & (50) \\
\midrule
Dev.-Baseline  & 0.088 & 0.319 & 0.59 & 0.82 & 0.57 & 0.90 & 0.53 & 0.78  \\
Vanilla-3DGS  & 0.127 & 0.165 & 0.32 & 0.38 & 0.42 & 0.62 & 0.39 & 0.45  \\
Scaffold-GS   & 0.102 & 0.135 &  0.25 & 0.34 & 0.39 & 0.54  & 0.37 & 0.40  \\
Ours GS-only & \underline{0.064} & \underline{0.102} & \underline{0.20} & \underline{0.26} & \underline{0.33} & \underline{0.48} & \underline{0.32} & \underline{0.35} \\
Ours   & \textbf{0.058} & \textbf{0.087} & \textbf{0.19}  & \textbf{0.21} & \textbf{0.26} & \textbf{0.39} & \textbf{0.28} & \textbf{0.33} \\
\bottomrule
\end{tabular}
}
\vspace{-0.2cm}
\label{tab:metrics_utwr}
\end{table}

\subsection{Adapting Base GS to Unconstrained Scenes}



We also integrate several enhancements to our base GS pipeline to address challenges in modeling unconstrained scenes, focusing on color representation, depth modeling, dynamic objects, appearance variability, and efficiency.


\noindent \textbf{Softmax-Depth Loss with Monocular Depth Priors: }
To improve geometric reconstruction, we introduce a scale-invariant depth loss $\mathcal{L}_{\text{sD}}$ guided by monocular depth priors (Marigold~\cite{ke2024repurposing}). Instead of alpha-blending, we use softmax-scaled depth rendering~\cite{xiong2023sparsegs}, which emphasizes depth gradients by weighting depth values with Gaussian opacities, prioritizing solid objects and reducing floating artifacts.



\noindent \textbf{Appearance Embedding for Lighting Variability: }We adopt global appearance embedding~\cite{martin2021nerf}, to compensate for appearance and lighting variability in input images. Each training image is assigned a learnable embedding vector, which helps
adapt the GS model to in-the-wild scenes.



\noindent \textbf{Dynamic Object Handling with Masks: }To mitigate the impact of dynamic objects, we use binary masks (indicating selected objects) generated using a semantic segmentation model. This mask $\mathcal{M}$ is then used to multiply the per-pixel loss defined in Eq.1 and the training loss effectively ignores the dynamic object regions. 



\noindent \textbf{Rasterization Improvements: } 
We incorporate two recent improvements to the 3DGS Rasterizer. First, absolute pixel gradients~\cite{ye2024absgs, yu2024gaussian} replace directional gradients to better differentiate well-reconstructed regions. Second, Mip-Splatting~\cite{yu2024mip} mitigates aliasing artifacts by using a 2D low-pass Mip filter, replacing the screen-space dilation filter.

\noindent \textbf{Opacity and Scaling Regularization: } We also apply regularization~\cite{xie2024physgaussian} to opacity and scaling to encourage fewer Gaussians, improving the efficiency of 3D representation.









\begin{table}[htbp]
\centering
\renewcommand{\arraystretch}{1.05}
\caption{Performance Metrics for Various Methods on Photo Tourism Dataset. The \textbf{best} and \underline{second-best}
results are highlighted. }
\vspace{-0.1cm}
\resizebox{\linewidth}{!}{%
\begin{tabular}{l|cc|cc|cc}
\toprule
\multirow{2}{*}{Method} & \multicolumn{2}{c|}{Brandenburg Gate} & \multicolumn{2}{c|}{Sacre Coeur} & \multicolumn{2}{c}{Trevi Fountain} \\
\cmidrule(lr){2-3} \cmidrule(lr){4-5} \cmidrule(lr){6-7}
 & PSNR↑ & SSIM↑ & PSNR↑ & SSIM↑ & PSNR↑ & SSIM↑ \\
\midrule
NeRF~\cite{mildenhall2021nerf} & 18.90 & 0.815 & 15.60 & 0.715 & 16.14 & 0.600 \\
NeRF-W~\cite{martin2021nerf} & 24.17 & 0.890 & 19.20 & 0.807 & 18.97 & 0.698 \\
Ref-Fields~\cite{kassab2023refinedfields} & 26.64 & 0.886 & 22.26 & 0.817 & 23.42 & 0.737 \\
3DGS~\cite{kerbl20233d} & 19.37 & 0.880 & 17.44 & 0.835 & 17.58 & 0.709 \\
GS-W~\cite{zhang2025gaussian} & 23.51 & 0.897 & 19.39 & 0.825 & 20.06 & 0.723 \\
SWAG~\cite{dahmani2025swag} & 26.33 & \textbf{0.929} & 21.16 & \underline{0.860} & 23.10 & \textbf{0.815} \\
WildGS~\cite{kulhanek2024wildgaussians} & \underline{27.77} & \underline{0.927} & \underline{22.56} & 0.859 & \textbf{23.63} & 0.766 \\
Ours & \textbf{28.69} & \textbf{0.929} & \textbf{23.76} & \textbf{0.861} & \underline{23.35} & \underline{0.767} \\
\bottomrule
\end{tabular}
}
\vspace{-0.1cm}
\label{tab:metrics_pt}
\end{table}

\begin{table}[ht]
\centering
\renewcommand{\arraystretch}{1.05}
\caption{Results of several GS Methods on Nerf On-the-go Sets.
\vspace{-0.15cm}
}
\label{tab:my_table}
\resizebox{\linewidth}{!}{%
\begin{tabular}{l|cc|cc|cc}
\toprule
Method & \multicolumn{2}{c|}{Low Occlusion} & \multicolumn{2}{c|}{Medium Occlusion} & \multicolumn{2}{c}{High Occlusion} \\
\cmidrule(lr){2-3} \cmidrule(lr){4-5} \cmidrule(lr){6-7}
 & PSNR$\uparrow$ & SSIM$\uparrow$ & PSNR$\uparrow$ & SSIM$\uparrow$ & PSNR$\uparrow$ & SSIM$\uparrow$ \\
\midrule
3DGS~\cite{kerbl20233d} & 19.68 & 0.649 & 19.19 & 0.709 & 19.03 & 0.649 \\
GS-Opacity~\cite{yu2024gaussian} & \underline{20.54} & \underline{0.662} & 19.39 & 0.719 & 17.81 & 0.578 \\
Mip-Splat~\cite{yu2024mip} & 20.15 & 0.661 & 19.12 & 0.719 & 18.10 & 0.664 \\
GS-W~\cite{zhang2025gaussian} & 18.67 & 0.595 & 21.50 & 0.783 & 18.52 & 0.644 \\
WildGS~\cite{kulhanek2024wildgaussians} & \textbf{20.62} & 0.658 & \textbf{22.80} & \underline{0.811} & \textbf{23.03} & \underline{0.771} \\
Ours & 19.85 & \textbf{0.777} & \underline{21.91} & \textbf{0.822} & \underline{21.65} & \textbf{0.832} \\
\bottomrule
\end{tabular}
}
\vspace{-0.2cm}
\label{tab:metrics_ngo}
\end{table}

\section{Experiments}



\noindent \textbf{Implementation Details: } Our models are trained on a NVIDIA A5000 GPU with $\lambda_{ssim}$, $\lambda_{GS}$, and $\lambda_{sD}$ set to 0.2, 0.5, and 0.1 respectively. The MLPs follow the Scaffold-GS framework~\cite{lu2024scaffold}. View augmentation is applied every 3rd training iteration. The multi-view diffusion model is trained on the DL3DV-10K~\cite{ling2024dl3dv} dataset.

\noindent \textbf{Datasets: } 
We conduct our experiments on four datasets. The Photo Tourism dataset features unconstrained user-uploaded images of popular monuments captured at different times and with varying camera models~\cite{snavely2006photo}. The NeRF On-the-go dataset includes casually captured indoor and outdoor sequences with varying occlusions~\cite{ren2024nerf}. We also evaluate on two WRIVA public data sets~\cite{cjk5-gf33-24}—APL Image Density iPAD (WRIVA-AIDI) and Massachusetts TaskForce Artifact (WRIVA-MTA)—which differ in image density and artifacts. Additionally, we test on two ULTRRA Development Sets~\cite{2zs6-ht63-24}, i.e., ImageDensity t01-v09-s00-r01 (ID-1901) and CameraModels t02-v06-s00-r01 (CM-2601).

\subsection{Results Analysis}
Tab.~\ref{tab:metrics_utwr}, Tab.~\ref{tab:metrics_pt} and Tab.~\ref{tab:metrics_ngo} compares the proposed GS-Diff approach against several baselines and state-of-the-art methods. From Tab.~\ref{tab:metrics_utwr}, we observe that for a dense view set with good coverage in small area (e.g., ULTRRA ID-1901), all baselines perform comparably. However, in the WRIVA-AIDI experiments with sparse coverage, our method demonstrates a significant performance advantage. This highlights that integrating the diffusion prior and enhancing the base GS leads to notable performance improvements, particularly in sparse-view synthesis. For instance, in S06 set with $15$ images, DreamSim drops from $0.34$ for the base Scaffold-GS to $0.26$ for our GS-only model, and further improves to $0.21$ with the addition of diffusion prior. We also see significant improvement compared to the development baseline (i.e., Nerfacto) and vanilla 3DGS.


Experiments on the Photo Tourism dataset (Table~\ref{tab:metrics_pt}) demonstrate that our approach performs comparably to the best prior methods in managing significant appearance variations across inputs, while achieving overall better results than the state-of-the-art methods WildGS and SWAG. This highlights our method's ability to preserve sharp details while effectively handling appearance changes. Similarly, on the ULTRRA CM-2601 dataset, which includes multiple camera models and appearance variations, our approach again outperforms the baselines.

Our experiments on the NeRF On-the-go dataset (Tab.~\ref{tab:metrics_ngo}) show that our model performs on par with prior methods in handling transient objects within scenes. However, since our approach assumes a fixed set of object classes for transients, it may struggle when an unknown object class appears as an occluder in the scene. Furthermore, results on the WRIVA-MTA sets in Table~\ref{tab:metrics_utwr} show that our approach performs reasonably well in handling artifacts within the scene, despite not incorporating a dedicated artifact-handling mechanism in our GS framework.

\begin{figure}[t]
  \centering
\includegraphics[width=0.98\linewidth]{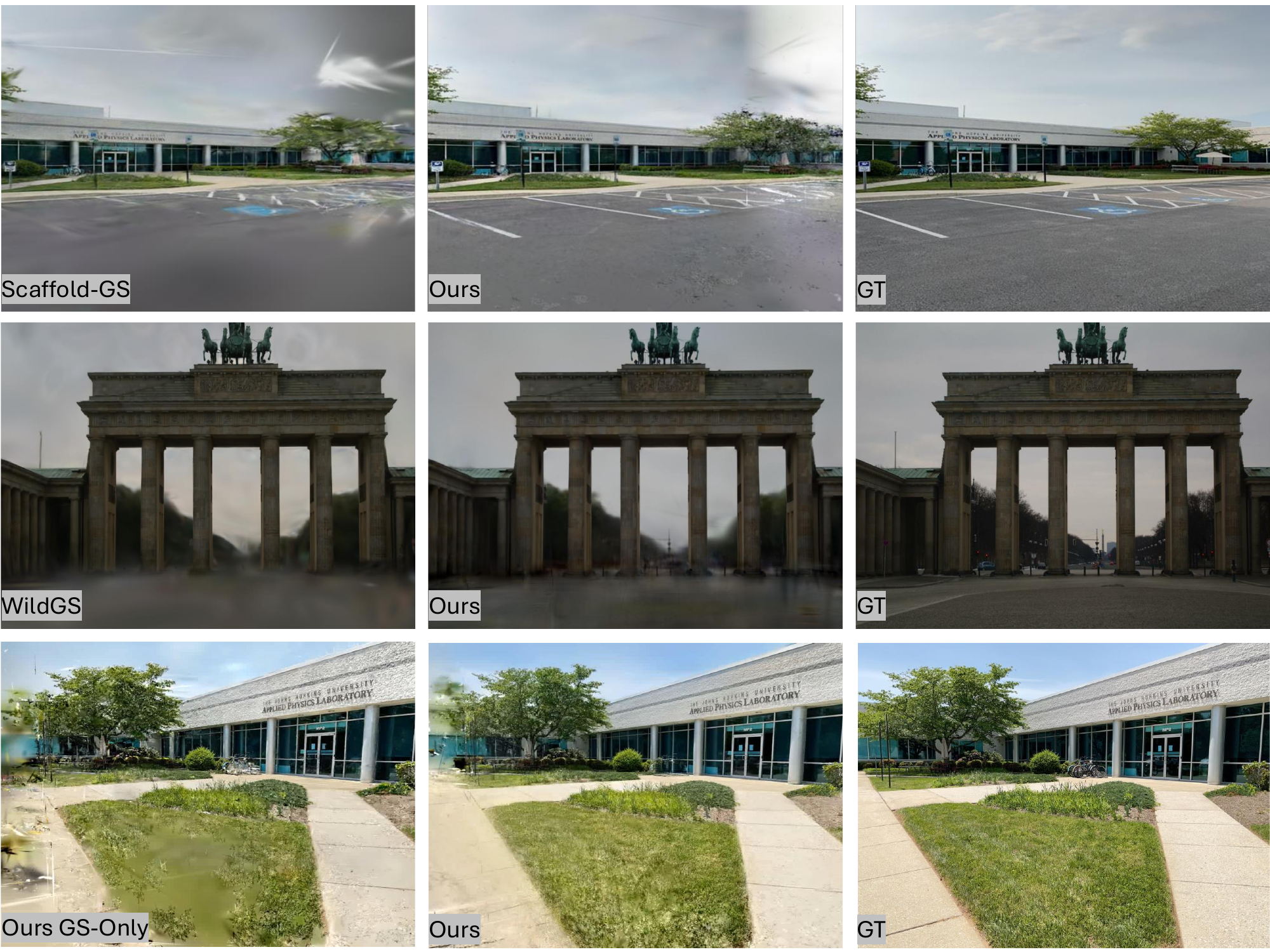}
  \caption{Comparison on the ULTRRA CM-2601 set (row-1), Photo Tourism Brandenburg Gate set (row-2), and WRIVA-AIDI 25 image set (row-3). Baselines (Left), Ours (Middle), GT (Right).}
  \label{fig:qualit}
\vspace{-0.4cm}
\end{figure}

Figure~\ref{fig:qualit} presents three qualitative examples comparing our method to the baselines. It is clear that our approach produces less artifacts than the baseline methods (e.g., row-1 and row-3). Additionally, our method recovers finer details; for instance, in row-2, compared to WildGS, we observe improved textures in the bottom part of the image, which was heavily occluded in the training set.

\section{Conclusions}
Our proposed GS-Diff presents a novel 3D reconstruction framework that integrates the efficiency of 3D Gaussian Splatting with the generalization capabilities of multi-view diffusion models, while incorporating critical enhancements to tackle in-the-wild challenges. Experiments across multiple benchmark datasets demonstrate the high-quality performance of our method in addressing various challenges in unconstrained 3D reconstruction.


\section{Acknowledgement}

Supported by the Intelligence Advanced Research Projects Activity (IARPA) via Department of Interior/ Interior Business Center (DOI/IBC) contract number 140D0423C0075. The U.S. Government is authorized to reproduce and distribute reprints for Governmental purposes notwithstanding any copyright annotation thereon. Disclaimer: The views and conclusions contained herein are those of the authors and should not be interpreted as necessarily representing the official policies or endorsements, either expressed or implied, of IARPA, DOI/IBC, or the U.S. Government.

{\small
\bibliographystyle{ieee_fullname}
\bibliography{egbib}
}

\end{document}